\documentclass[runningheads]{llncs}
\usepackage{graphicx}
\usepackage{amsfonts}
\usepackage{amssymb}
\usepackage{siunitx}
\usepackage{float}
\usepackage{color}
\usepackage{array, makecell}
\usepackage{caption}
\begin{document}
\title{Empirical Analysis of Machine Learning Configurations for Prediction of Multiple Organ Failure in  Trauma Patients}
\titlerunning{ML on MOF Prediction}
%
\author{Yuqing Wang\inst{1\star} \and
Yun Zhao\inst{1\star} \and
Rachael Callcut\inst{2} \and
Linda Petzold\inst{1}}

\renewcommand{\thefootnote}{\fnsymbol{footnote}}
\footnotetext[1]{These two authors contributed  equally to this paper.}

\authorrunning{Y. Wang, Y. Zhao et al.}
%
\institute{Department of Computer Science, University of California, Santa Barbara \and
UC, Davis Health\\
\email{wang603@ucsb.edu, yunzhao@cs.ucsb.edu}\\
}
\maketitle              
\begin{abstract}
Multiple organ failure (MOF) is a life-threatening condition. Due to its urgency and high mortality rate, early detection is critical for clinicians to provide appropriate treatment. In this paper, we perform quantitative analysis on early MOF prediction with comprehensive machine learning (ML) configurations, including data preprocessing (missing value treatment, label balancing, feature scaling), feature selection, classifier choice, and hyperparameter tuning. Results show that classifier choice impacts both the performance improvement and variation most among all the configurations. In general, complex classifiers including ensemble methods can provide better performance than simple classifiers. However, blindly pursuing complex classifiers is unwise as it also brings the risk of greater performance variation. 

\end{abstract}

\section{Introduction}

Multiple organ failure (MOF) is a clinical syndrome with variable causes including pathogens~\cite{Harjola}, complicated pathogenesis~\cite{Wang}, and a major cause of mortality and morbidity for trauma patients who are admitted to Intensive Care Units (ICU)~\cite{Durham}. Based on recent studies on ICU trauma patients, up to $47\%$ have developed MOF, and MOF increased the overall risk of death $6$ times compared to patients without MOF~\cite{Ulvik}. To prevent the development of MOF for trauma patients from progression to an irreversible stage, it is essential to diagnose MOF early and effectively. Many scoring systems have been proposed to predict MOF~\cite{Barie,Bota,Dewar,Hutchings} and researchers have attempted to predict MOF on trauma patients using predictive models in an early phase~\cite{Sauaia,Vogel}.  

The rapid growth of data availability in clinical medicine requires doctors to handle extensive amounts of data. As medical technologies become more complicated, technological advances like machine learning (ML) are increasingly needed to improve real-time analysis and interpretation of the results~\cite{Obermeyer}. In recent years,  practical uses of ML in healthcare have grown tremendously, 
including cancer diagnosis and prediction~\cite{Cruz,Kourou,Asri}, tumor detection~\cite{Sharma,Zhiqiong}, medical image analysis~\cite{De}, and health monitoring~\cite{Farrar,Worden}.  

Compared to traditional medical care, ML-assisted clinical decision support enables a more standardized process for interpreting complex multi-modality data. In the long term, ML can provide an objective viewpoint for clinical practitioners to improve performance and efficiency~\cite{Ahmed}. ML is often referred to as a black box: explicit input data and output decisions, but opaque at intermediate learning process. Additionally, in medical domains, there is no universal rule for selecting the best configuration to achieve the optimal outcome. Moreover, medical data has its own challenges such as numerous missing values~\cite{Janssen} and colinear variables~\cite{Tuba}. Thus it is difficult to process the data and choose the proper
model and corresponding parameters, even for a ML expert. Furthermore, detailed quantitative analysis of the potential impacts of different settings of ML systems on MOF has been missing.

In this paper, we experiment with comprehensive ML settings for prediction of MOF, considering $6$ different dimensions from data preprocessing (missing value treatment, label balancing, feature scaling), feature selection, classifier choice, to hyperparameter tuning. To predict MOF
for trauma patients at an early stage, we  use only initial time measurements (hour $0$) as inputs. 
We mainly use area under the receiver operating characteristic curve (AUC) to evaluate MOF prediction outcomes. We focus on analyzing the relationships among configuration complexity, predicted performance, and performance variation. Additionally, we quantify the relative impacts of each dimension.

The main contributions of this paper include:
\begin{enumerate}
    \item [(1)] To the best of our knowledge, this is the first paper to conduct a thorough empirical analysis quantifying the predictive performance with exhaustive ML configurations for MOF prediction.
    \item [(2)] We provide general guidance for ML practitioners in healthcare and medical fields through quantitative analysis of different dimensions commonly used in ML tasks.
    \item [(3)] Experimental results indicate that classifier choice contributes most to both performance improvement and variation. Complex classifiers including ensemble methods bring higher default/optimized performance, along with a higher risk of inferior performance compared to simple ones on average.
\end{enumerate}
The remainder of this paper is organized as follows. Section~\ref{dataset} describes the dataset and features we use. All of the ML configurations are available in Section~\ref{methods}. Experimental results are discussed in Section~\ref{results}. Finally, our conclusions are presented in Section~\ref{discussion}. 

\section{Dataset}\label{dataset}
Our dataset, collected from the San Francisco General Hospital and Trauma Center, contains $2190$ highest level trauma activation patients evaluated at the level I trauma center. Due to the urgency of medical treatment, there are numerous missing values for time-dependent measurements. Thus we have chosen to consider only those features with a maximum missing value percentage of $30\%$ over all patients. To obtain a timely prediction, early lab measurements (hour $0$) as well as patients' demographic and illness information were extracted as the set of features. 
Detailed feature 
statistics are available in Table~\ref{table_dataset}.

\begin{table}[H]
\centering
\begin{tabular}{|c|c|c|}\hline
    \ Feature type \ & \ \makecell{\#  of \\ extracted \\ features} \ & Features \ \\ \hline
	\ Demographic \  & $5$ & \ \textit{gender}, age, weight, \textit{race, blood type} \ \\ \hline
	\ Illness \ & $2$ & \ \textit{comorbidities, drug usage} \ \\ \hline
	\ Injury factors \ & $4$ & \ \makecell{\textit{blunt/penetrating trauma}, \\ \# of rib fractures, \\ \textit{orthopedic injury}, \textit{traumatic brain injury}} \ \\ \hline
	\ Injury scores \ & $8$ & \ \makecell{injury severity score, \\ $6$ abbreviated injury scale (head, face, chest, \\ abdomen, extremity, skin), \\ Glasgow coma scale score} \\ \hline
	\makecell{Vital sign \\ measurements} & $4$ & \ \makecell{heart rate, respiratory rate, \\ systolic blood pressure, \\ mean arterial pressure} \\ \hline
	\makecell{Blood-related \\ measurements} & $13$ & \makecell{white blood cell count, \\ hemoglobin, hematocrit, \\ serum CO\textsubscript{2}, 
	prothrombin time, \\ international normalized ratio, \\ partial thromboplastin time, \\ blood urine nitrogen, creatinine, \\ blood pH, platelets, base deficit, \\ \textit{factor VII}} \\ \hline

\end{tabular}
\caption{MOF dataset statistics. Italicized features are categorical.} \label{table_dataset}
\end{table}
\vspace{-1cm}
Our target variable consists of binary class labels ($0$ for no MOF and $1$ for MOF). Then, the data with feature and target variables is randomly split into training and testing sets at the ratio of $7 : 3$. 

\section{Methods}\label{methods}
Based on ML pipelines and special characteristics of our data such as large number of missing values and varying scales in feature values, we consider comprehensive ML configurations from the following $6$ dimensions: data preprocessing (missing value treatment (MV), label balancing (LB), feature scaling (SCALE)), feature selection (FS), classifier choice (CC), and hyperparameter tuning (HT). In the remainder of the paper, we will interchangeably use the full name and corresponding abbreviations shown in parentheses. Further details on each dimension are 
described below.
\subsection{Data Preprocessing}
Methods to handle the dataset with missing values, imbalanced labels, and unscaled variables are essential for the data preprocessing process. We use several different methods to deal with each of these problems.
\subsubsection{Missing Value Treatment}
In our dataset, numerous time-dependent features cannot be recorded on a timely basis, and missing data is a serious issue. We consider three different ways to deal with missing values, where the first method serves as the baseline setting for MV, and the latter two methods are common techniques of missing value imputation in ML.
\begin{enumerate}
    \item Remove all patients with any missing values for the features listed in Section~\ref{dataset}. 
    \item Replace missing values with mean for numerical features and mode for categorical features over all patients.
    \item Impute missing values by finding the $k$-nearest neighbors with the Euclidean distance metric for each feature respectively.
\end{enumerate}
\subsubsection{Label Balancing}
Our dataset is imbalanced as the sample class ratio between class $0$ and class $1$ is $11 : 1$. 
Keeping imbalanced class labels serves as the baseline setting for LB. Three different ways are considered to resample the training set. 
\begin{enumerate}
    \item Oversampling the minority class (label $1$) 
    \begin{itemize}
        \item[1.1] Method: SMOTE (synthetic minority over-sampling technique)~\cite{Chawla}.
        \item[1.2] Explanation: choose $k$-nearest neighbors for every minority sample and then create new samples halfway between the original sample and its neighbors.
    \end{itemize}
    \item Undersampling the majority class (label $0$)
    \begin{itemize}
        \item[2.1] Method: NearMiss~\cite{Mani}.
        \item[2.2] Explanation: when samples of both classes are close to each other, remove the
    samples of the majority class to provide more space for both classes.
    \end{itemize}
        \item Combination of oversampling and undersampling
        \begin{itemize}
            \item[3.1] Method: SMOTE \& Tomek link~\cite{Batista}.
            \item[3.2] Tomek link: two samples are $k$-nearest neighbors to each other but come from different classes.
            \item[3.3] Explanation: first create new samples for the minority class and then remove the majority class sample in any Tomek link.
        \end{itemize}
\end{enumerate}
\subsubsection{Feature Scaling} Since the range of feature values in our dataset varies widely, we perform feature scaling. No scaling on any feature serves as the baseline setting for SCALE. Two common scaling techniques are used for numerical features.
\begin{enumerate}
    \item Normalization: rescale values to range between $0$ and $1$.
    \item Standardization: rescale values with mean $0$ and standard deviation $1$.
\end{enumerate}

\subsection{Feature Selection}

In medical datasets, there usually exist many highly correlated features, and some features that are weakly correlated to the target~\cite{Tuba,Dag}. Thus it is essential to identify the most relevant features that may help to improve the outcome of the analysis.  Using all of the features described in Section $2$ serves as the baseline setting for FS. We consider two main feature selection techniques: filter and wrapper methods.

\begin{enumerate}
    \item Filter-based methods (independent of classifiers):
    \begin{itemize}
        \item[1.1] Use correlation between features and the target to select features which are highly dependent on the target.
        \item[1.2] Filter out numerical features using ANOVA $F$-test and categorical features using $\chi^2$ test. 
    \end{itemize}
    \item Wrapper-based methods (dependent on classifiers):
    \begin{itemize}
        \item[2.1] Method: RFE (recursive feature elimination) in random forest.
        \item[2.2] Explanation: perform RFE repeatedly such that features are ranked by importance, and the least important features are disregarded until a specific number of features remains.
    \end{itemize}
\end{enumerate}

\vspace{-.5cm}

\subsection{Classifier Choice}
We experimented with $15$ classifiers on the dataset. In general, these classifiers can be divided into two main categories: single and ensemble. Lists of all classifiers are available in Table~\ref{table_cc}. For ensemble classifiers (combination of individual classifiers), we tried bagging (BAG, RF, ET), boosting (GB, ABC, XGB, LGBM), voting (VOTE) and stacking (STACK). In bagging, DT is a homogeneous weak learner. Multiple DTs learn the dataset independently from each other in parallel and the final outcome is obtained by averaging the results of each DT. In boosting, DT also serves as a homogeneous weak learner. However, DTs learn the dataset sequentially in an adaptive way (new learner depends on previous learners' success), and the final outcome is determined by weighted sum of previous learners. In voting, heterogeneous base estimators (LR, RF, SVM, MLP, ET) are considered, where each estimator learns the original dataset and the final prediction is determined by majority voting. In stacking, several heterogeneous base learners (RF, KNN, SVM) learn the dataset in parallel, 
and there exists a meta learner (LR) that combines the predictions of the weak learners.
Abbreviations of classifiers shown in parentheses for voting and stacking are the ones we use.

\newpage
\begin{table}[H]
\centering
\setlength{\extrarowheight}{-1pt}
\begin{tabular}{|c|c|}\hline
\ Single classifiers \ & \ Ensemble classifiers \ \\ \hline
\ \makecell{Logistic Regression (LR) \\ Support Vector Machine (SVM) \\ Naive Bayes (NB) \\ K-nearest Neighbors (KNN) \\ Decision Tree (DT) \\ Multi-layer Perceptron (MLP)} \ & \ \makecell{Bagged Trees (BAG) \\ Random Forest (RF) \\ Extra Trees (ET) \\ Gradient Boosting (GB) \\ Adaptive Boosting (ABC) \\ Extreme Gradient Boosting (XGB) \\ Light Gradient Boosting Machine (LGBM) \\ Voting (VOTE) \\ Stacking (STACK)} \\ \hline
\end{tabular}
\caption{List of $6$ single classifiers and $9$  ensemble classifiers. Corresponding abbreviations of each classifier are shown in parentheses.} \label{table_cc}
\end{table}

\vspace{-1cm}
\subsection{Hyperparameter Tuning}
Hyperparameters are crucial for controlling the overall behavior of classifiers. Default hyperparameters of classifiers serve as the baseline setting for HT. We apply grid search to perform hyperparameter tuning for all classifiers. Detailed information about tuned hyperparameters is available in Table~\ref{table_ht}.
\vspace{-0.5cm}
\begin{table}[H]
\centering
\begin{tabular}{|c|c|c|c|c|}\hline
	 \ Classifiers \ & \ \makecell{$\sharp$ of tuned \\ hyperparameters} \ & \ Hyperparameter lists \  \\ \hline
	 LR & $3$ & C, class\_weight, penalty \\ \hline
	 SVM & $4$ & $C$, gamma, kernel, class\_weight \\ \hline
	 KNN & $3$ & n\_neighbors, weights, algorithm \\ \hline
	 NB & $1$ & var\_smoothing \\ \hline 
	 DT & $5$ & \makecell{min\_samples\_split, max\_depth,  min\_samples, \\ leaf\_max\_features, class\_weight} \\ \hline
	 MLP & $3$ & activation, solver, alpha \\ \hline
	 BAG & $2$ & base\_estimator, n\_estimators \\ \hline
	 RF & $2$ & n\_estimators, max\_features \\ \hline
	 ET & $2$ & n\_estimators, max\_features \\ \hline
	 GB & $2$ & n\_estimators, max\_depth \\ \hline
	 ABC & $3$ & base\_estimator, n\_estimators, learning\_rate \\ \hline
	 XGB & $2$ & min\_child\_weight, max\_depth \\ \hline
	 LGBM & $4$ & num\_leaves, colsample\_bytree, subsample, max\_depth \\ \hline 
	 VOTE & $2$ & C (SVM), n\_estimators (ET) \\ \hline
	 STACK & $2$ & C (SVM), n\_neighbors (KNN) \\ \hline
\end{tabular}
\caption{Detailed configurations of tuned hyperparameters for all classifiers. All of the hyperparameter names come from \textit{scikit-learn}~\cite{Pedregosa}.} \label{table_ht}
\end{table}

\newpage
\section{Experiments and Results}\label{results}

We formulated MOF prediction as a binary classification task. All of the experiments in this paper
were implemented using \textit{scikit-learn}~\cite{Pedregosa}.
As mentioned in Section~\ref{dataset}, our training and testing dataset is randomly split with a ratio of $7:3$. One-hot encoding is applied to all categorical features. For each classifier, we use the same training and testing dataset. We use AUC as our main performance metric, as it is commonly used for MOF prediction in the literature~\cite{Bota,Bakker,Papachristou}. It provides a “summary" of classifier performance compared to single metrics such as precision and recall. AUC represents the probability that a classifier ranks a randomly chosen positive sample (class $1$) higher than a randomly chosen negative sample (class $0$), and thus useful for imbalanced datasets.
In this section, we quantify the 
impacts (improvement and variation) of each dimension on the predicted performance over our testing dataset.

\subsection{Influence of Individual Dimensions}
First, we evaluate how much each dimension contributes to the AUC score improvement and variation respectively, and find the correlation between performance improvement and variation over all dimensions. 
\subsubsection{Performance Improvement across Dimensions} 

\begin{figure}[ht]
\centering
\includegraphics[width=0.9\linewidth]{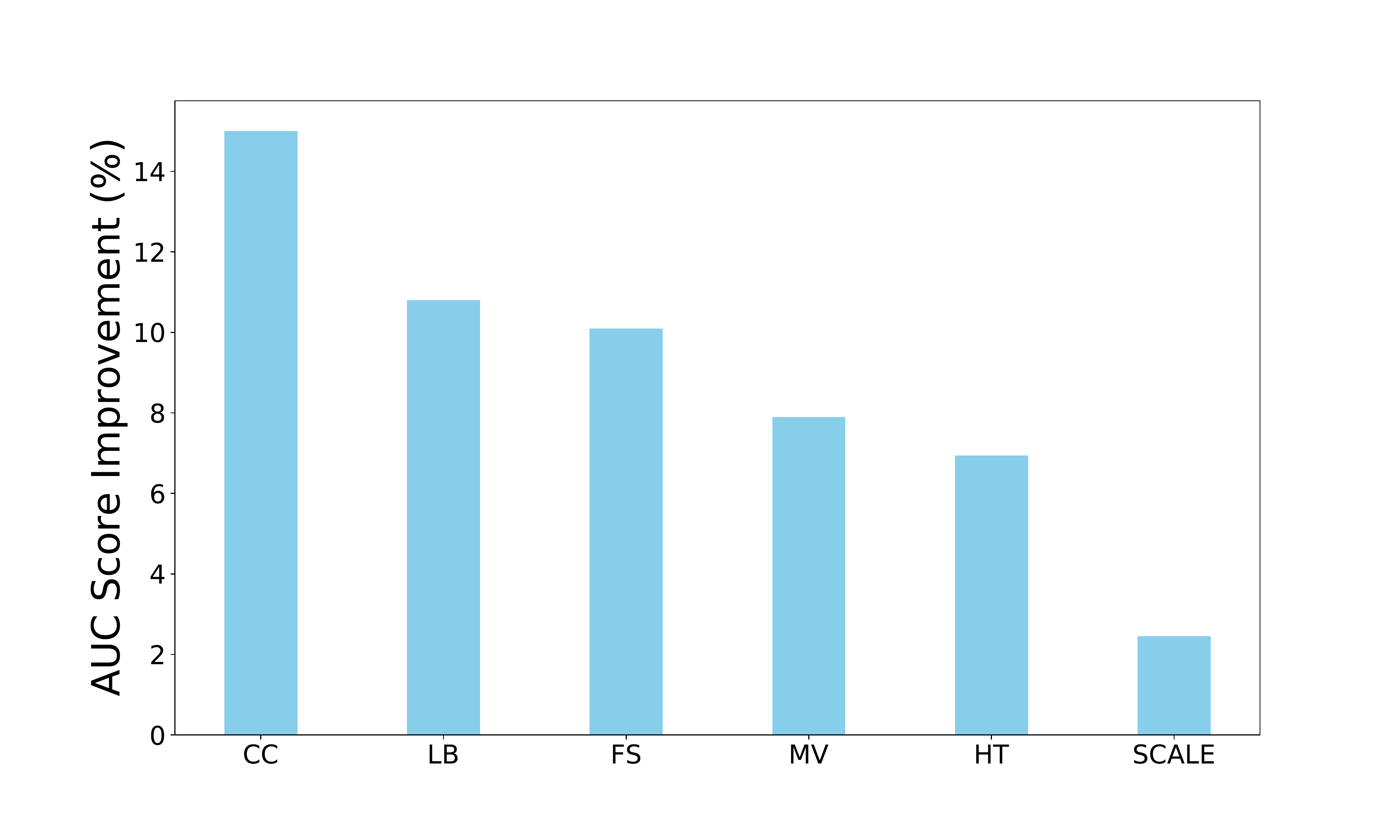}
\caption{Performance improvement in the AUC score of each dimension over the baseline when tuning only one dimension at a time while leaving others at baseline settings. CC brings the greatest performance improvement, followed by LB, FS, MV, HT, and SCALE in decreasing order of improvement.} \label{fig_comp}
\end{figure}

For HT, MV, LB, SCALE, and FS, we define the $\textbf{\emph{baseline}}$ as default hyperparameter choices, using no missing value imputation, no label balancing, no feature scaling, and no feature selection, respectively. For CC, we choose SVM, which achieves the median score among all classifiers, as the $\textbf{\emph{baseline}}$.
Then we quantify the performance improvement of each dimension. Fig.~\ref{fig_comp} shows the percentage that each dimension contributes to the improvement in the AUC score over baseline by tuning only one dimension at a time while leaving others at baseline settings.
We observe that CC contributes most to the performance improvement ($15.00 \%$) for MOF prediction. After CC, LB ($10.81 \%$), FS ($10.09 \%$), MV ($7.90 \%$), HT ($6.94 \%$), and FS ($2.45 \%$) 
bring decreasing degrees of performance improvement in the AUC score.

Table~\ref{table_comp} shows the improvement of every single dimension on each classifier over the baseline. In general, MV and LB tend to provide the greatest performance improvement for most classifiers. For RF, ET, and LGBM, FS contributes the most to improvement in performance since these classifiers require feature importance ranking intrinsically, and external FS improves their prediction outcomes to a large extent. 
Note that the classifier for which SCALE has the largest impact is KNN, as it is a distance-based classifier which is sensitive to the range of feature values. Also, due to instability and tendency to overfit, HT is the most critical for DT improvement.

\begin{table}[H]
\centering
\begin{tabular}{|c|c|c|c|c|c|}\hline
	\ Classifier \ & \ MV (\%) \ & \ LB (\%) \ & \ SCALE (\%) \ & \ FS (\%) \ & \ HT (\%) \ \\ \hline
	LR & $2.78$ & $\textbf{11.48}$ & $0.30$ & $5.50$ & $3.03$\\ \hline
	SVM & $3.37$ & $\textbf{26.83}$ & $2.38$ & $20.95$ & $3.21$\\ \hline
	KNN & $13.60$ & $11.85$ & $\textbf{17.68}$ & $13.12$ & $15.60$ \\ \hline
	NB & $0.60$ & $\textbf{38.90}$ & $0.17$ & $4.12$ & $2.84$ \\\hline
	DT & $12.87$ & $16.22$ & $0.42$ & $15.34$ & $\textbf{38.85}$\\ \hline
	BAG & $2.94$ & $\textbf{8.91}$ & $0.28$ & $7.05$ & $5.43$ \\ \hline
	RF & $4.13$ & $5.34$ & $0.28$ & $\textbf{5.85}$ & $1.04$ \\ \hline
	ET & $3.82$ & $7.87$ & $0.00$ & $\textbf{18.96}$ & $1.33$ \\ \hline
	ABC & $\textbf{19.33}$ & $7.02$ & $0.00$ & $16.99$ & $12.99$ \\ \hline
	GB & $\textbf{12.44}$ & $3.81$ & $0.02$ & $6.63$ & $4.08$ \\ \hline
	LGBM & $7.03$ & $1.85$ & $2.75$ & $\textbf{10.39}$ & $3.13$ \\ \hline
	XGB & $\textbf{11.46}$ & $3.97$ & $0.02$ & $7.47$ & $4.27$ \\ \hline
	MLP & $\textbf{10.78}$ & $5.08$ & $6.05$ & $7.53$ & $5.69$ \\ \hline
	STACK & $6.94$ & $\textbf{8.94}$ & $4.32$ & $5.48$ & $1.82$ \\ \hline
	VOTE & $\textbf{6.38}$ & $4.00$ & $2.11$ & $6.04$ & $0.85$ \\ \hline
\end{tabular}
\caption{Column $1$ shows a total of $15$ classifiers. Columns $2$ to $6$ represent the percentage (two decimal places accuracy) of AUC score improvement when tuning each individual dimension while leaving other dimensions at baseline settings for each classifier. Bold entries represent the dimension that contributes to the largest improvement for the specific classifier. MV and LB 
tend to dominate in performance improvement for most classifiers.}  
\label{table_comp}
\end{table}

In addition to AUC, $6$ other performance metrics are used to measure the performance improvement degree of each dimension. The results in  Table~\ref{table_metrics_dim_imp} reveal that CC brings the greatest improvement regardless of the metrics we use. 
Contributions from HT and SCALE are relatively small compared to other dimensions.

\vspace{-.5cm}
\begin{table}[H]
\centering
\setlength{\tabcolsep}{.4pt}
\begin{tabular}{|c|c|c|c|c|c|c|c|}\hline
    \ \	& \ AUC \ & \ F-score \  & \ G-mean \ & \ Precision \ & \ \makecell{Sensitivity/ \\ Recall} \ & \ Specificity \ & \ Accuracy \ \\ \hline
	CC (\%)  & $15.00$ & $15.58$ & $10.50$ & $16.41$ & $10.50$ & $11.86$ & $10.50$ \\ \hline
	 LB (\%)  & $10.81$ & $11.34$ & $9.33$ & $13.27$ & $9.33$ & $10.72$ & $9.34$ \\ \hline
	 FS (\%)  & $10.09$ & $7.33$ & $6.30$ & $10.61$ & $6.30$ & $6.94$ & $6.30$ \\ \hline
	MV (\%)  & $7.90$ & $5.30$  & $4.60$ & $5.83$ & $4.59$ & $4.95$ & $4.59$ \\ \hline
	 HT (\%)  & $7.46$ & $2.11$  & $3.21$ & $3.41$ & $3.20$ & $4.64$ & $3.20$\\ \hline
     SCALE (\%)  & $2.45$ & $1.04$ & $0.65$ & $3.03$ & $0.65$ & $0.48$ & $0.65$ \\ \hline
\end{tabular}
\caption{Performance improvement in different metrics of each dimension. The performance improvement of each dimension on other metrics displays an order consistent with that of the AUC score.} \label{table_metrics_dim_imp}
\end{table}

\subsubsection{Performance Variation across Dimensions}
\begin{figure}[ht]
\centering
\includegraphics[width=0.9\linewidth]{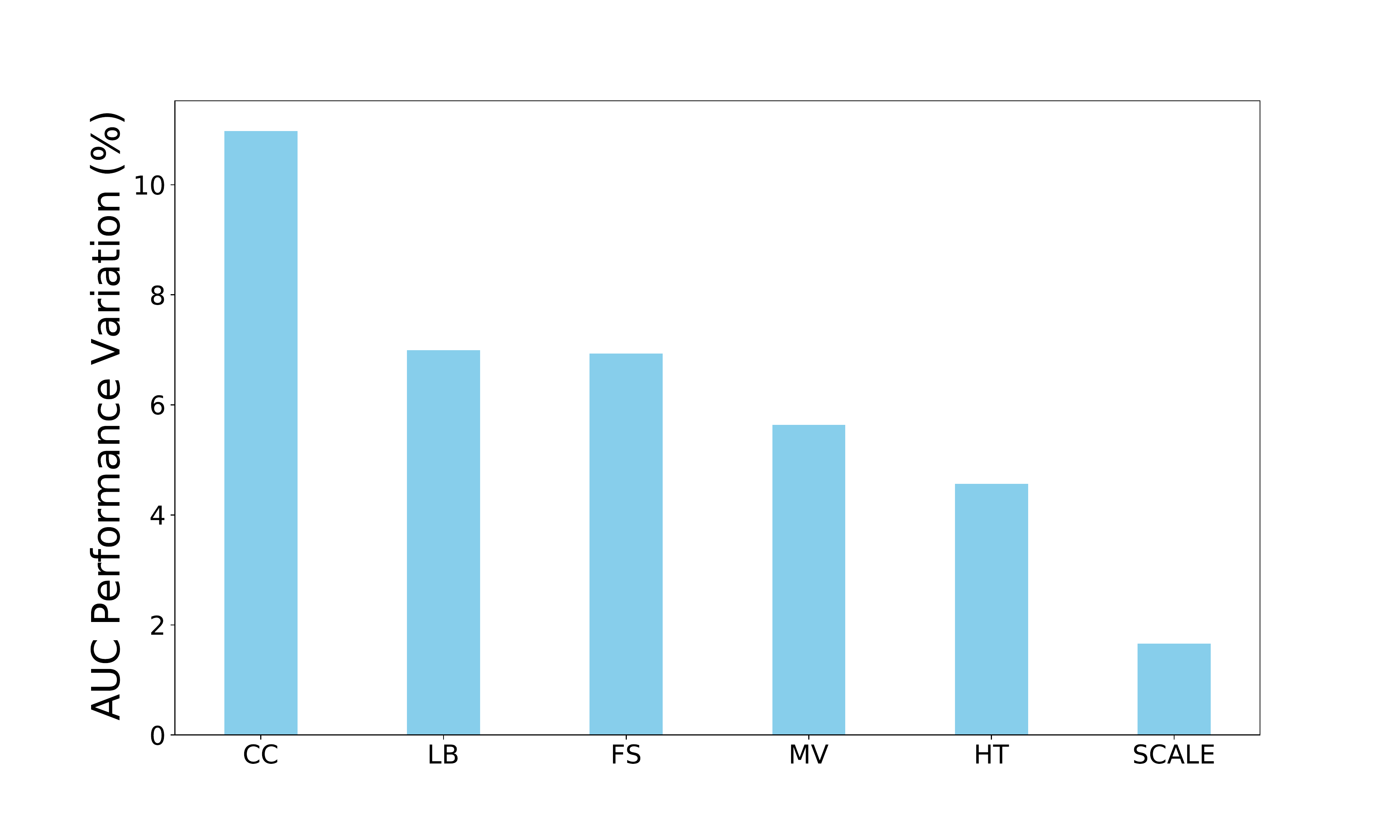}
\caption{Performance variation in the AUC score when tuning only one dimension at a time while leaving others at baseline settings. CC brings the greatest performance variation, followed by LB, FS, MV, HT, and SCALE in decreasing order of variation. Larger improvement also brings the risk of larger variation for each dimension.} \label{fig_var}
\end{figure}

For all of the ML configurations, we further investigate how much each dimension contributes to the performance variation in the AUC score. By tuning only one dimension at a time while leaving other dimensions at baseline settings, we obtain a range of AUC scores. Performance variation is the difference between the maximum and the minimum score of each dimension. Fig.~\ref{fig_var} shows the proportion of each dimension that brings the performance variation in the AUC score. Based on Fig.~\ref{fig_var}, we notice that CC, which brings the largest performance improvement, also brings the largest performance variation ($10.98$ \%). After CC, LB ($7.00$ \%), FS ($6.93$ \%), MV ($5.64$ \%), HT ($4.97$ \%), and SCALE ($1.66$ \%) bring decreasing degrees of performance variation in the AUC score.

Table~\ref{table_var} shows the variation of every single dimension on each classifier over the baseline. We observe that for each classifier, if one dimension brings a larger performance improvement, it also results in a larger performance variation. 
For our assessment of performance variation, the same metrics as above are used for evaluation on each dimension. Using the same metrics as above, Table~\ref{table_metrics_dim_var} shows that the proportion of performance variation in different metrics from each dimension follows an order that is consistent with the performance improvement in Table~\ref{table_metrics_dim_imp}. Thus, for different metrics, greater improvement brings greater variation of each dimension. For every step that researchers take when predicting MOF using ML, they should always be aware of the 
trade-off between benefits (improvement in performance) and risks (variation in performance) when adjusting each dimension. 

\begin{table}[H]
\centering
\begin{tabular}{|c|c|c|c|c|c|}\hline
		\ Classifier \ & \ MV (\%) \ & \ LB (\%) \ & \ SCALE (\%) \ & \ FS (\%) \ & \ HT (\%) \ \\ \hline
	LR & $2.28$ & $\textbf{8.44}$ & $0.25$ & $4.27$ & $2.49$\\ \hline
	SVM & $2.45$ & $\textbf{16.87}$ & $1.79$ & $13.04$ & $2.42$\\ \hline
	KNN & $7.97$ & $6.95$ & $\textbf{10.36}$ & $7.69$ & $9.14$ \\ \hline
	NB & $0.48$ & $\textbf{22.22}$ & $0.13$ & $3.25$ & $2.25$ \\\hline
	DT & $7.13$ & $8.99$ & $0.23$ & $8.50$ & $\textbf{21.53}$\\ \hline
	BAG & $2.22$ & $\textbf{6.17}$ & $0.21$ & $5.26$ & $4.10$ \\ \hline
	RF & $3.35$ & $4.14$ & $0.23$ & $\textbf{4.49}$ & $0.84$ \\ \hline
	ET & $3.22$ & $6.14$ & $0.00$ & $\textbf{13.41}$ & $1.12$ \\ \hline
	ABC & $\textbf{13.09}$ & $4.61$ & $0.00$ & $11.51$ & $9.40$ \\ \hline
	GB & $\textbf{9.59}$ & $2.83$ & $0.02$ & $5.11$ & $3.14$ \\ \hline
	LGBM & $5.54$ & $1.43$ & $2.16$ & $\textbf{7.76}$ & $2.47$ \\ \hline
	XGB & $\textbf{8.70}$ & $3.01$ & $0.02$ & $5.59$ & $3.24$ \\ \hline
	MLP & $\textbf{7.89}$ & $3.55$ & $4.43$ & $5.30$ & $4.16$ \\ \hline
	STACK & $5.47$ & $\textbf{6.47}$ & $3.41$ & $4.20$ & $1.43$ \\ \hline
	VOTE & $\textbf{5.19}$ & $3.13$ & $1.71$ & $4.63$ & $0.69$ \\ \hline
\end{tabular}
\caption{Columns $2$ to $6$ represent the proportion (two decimal places accuracy) of each dimension that contributes to the performance variation in the AUC score. Bold entries represent the dimension that contributes to the largest variation for the specific classifier. MV and LB tend to result in larger performance variation for most classifiers.} \label{table_var}
\end{table}

\begin{table}[H]
\centering
\setlength{\tabcolsep}{.4pt}
\begin{tabular}{|c|c|c|c|c|c|c|c|}\hline
	\ \	& \ AUC \ & \ F-score \  & \ G-mean \ & \ Precision \ & \ \makecell{Sensitivity/ \\ Recall} \ & \ Specificity \ & \ Accuracy \ \\ \hline
	 CC (\%)  & $10.98$ & $11.87$ & $8.57$ & $12.86$ & $8.57$ & $10.60$ & $8.57$\\ \hline
	 LB (\%)  & $7.00$ & $7.29$ & $6.83$ & $9.02$ & $6.83$ & $10.14$ & $6.82$ \\ \hline
	 FS (\%)  & $6.93$ & $5.27$ & $4.38$ & $7.62$ & $4.37$ & $4.70$ & $4.38$ \\ \hline
	 MV (\%)  & $5.64$ & $4.36$  & $3.69$ & $4.77$ & $3.69$ & $2.98$ & $3.68$ \\ \hline
	 HT (\%)  & $4.87$ & $2.55$  & $3.52$ & $1.54$ & $3.52$ & $2.72$ & $3.53$\\ \hline
     SCALE (\%) & $1.66$ & $0.88$ & $0.57$ & $1.47$ & $0.57$ & $0.46$ & $0.57$ \\ \hline
\end{tabular}
\caption{Performance variations in different metrics of each dimension. The performance variation of each dimension on other metrics displays an order that is consistent with that of the AUC score.} \label{table_metrics_dim_var}
\end{table}
\subsection{Performance Comparison across Classifiers}
\vspace{-.8cm}
We have shown that classifier choice is the largest contributor to both performance improvement and variation in the AUC score. Hence, we further investigate the performance differences among classifiers. Specifically, we investigate the relationships among classifier complexity, performance, and performance variation. 

\subsubsection{Default versus Optimized Performance}
\begin{figure}[h]
\centering
\includegraphics[width=0.85\linewidth]{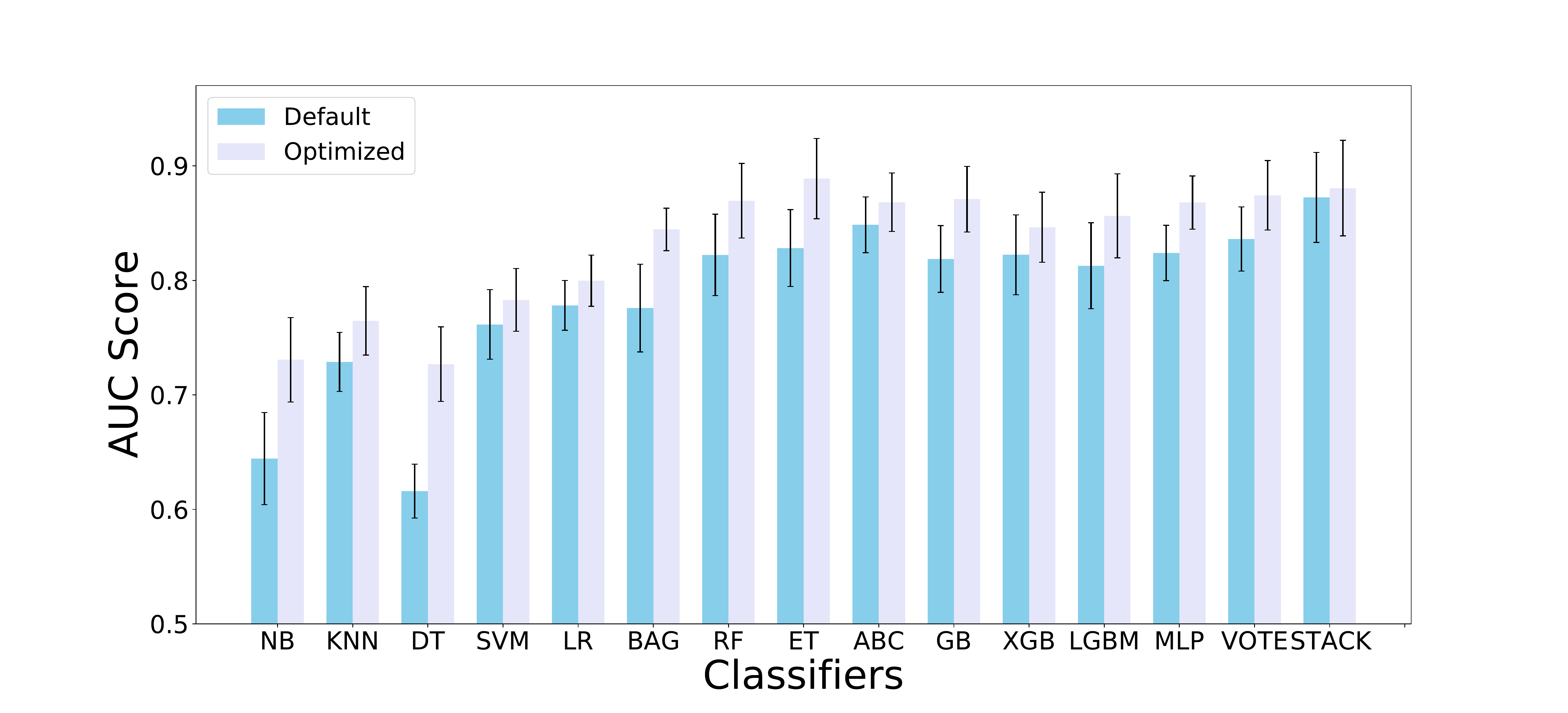}
\caption{Comparison of default and optimized performance over all classifiers. Classifiers listed on the left-hand side of BAG are single while the ones on the right-hand side are ensemble and MLP. Overall, ensemble methods have better default and optimized performance compared with single classifiers.} \label{fig_cccomp}
\end{figure}

\vspace{-1cm}
$\textbf{\emph{Default}}$ classifiers are defined as classifiers with default parameters, while $\textbf{\emph{optimized}}$ classifiers are those for which hyperparameter tuning with $10$-fold cross validation is applied using grid search. We compare the performance of default and optimized classifiers in consideration of all other dimensions, i.e., MV, LB, SCALE, and FS. The average AUC scores of all classifiers with default and optimized settings are 
shown in Fig.~\ref{fig_cccomp}. In general, ensemble classifiers perform better than 
single classifiers regardless of default or optimized performance. 

In addition to AUC, $6$ other performance metrics are used to evaluate the performance of all classifiers. We use the median score to rank classifiers with both default and optimized settings. Then, NDCG (normalized discounted cumulative gain), one of the most prevalent measures of ranking quality~\cite{Chen}, is used to compare classifier rankings between each of these metrics and the AUC score. Detailed relevance scores are shown in Table~\ref{table_ndcg}. The result indicates that the median performance of each classifier is similar no matter which metric 
is used. This also suggests that the AUC score can represent classifiers' overall performance well.

Based on the above experiments, ensemble classifiers should be prioritized in MOF prediction since they usually bring better predictive performance than single classifiers.

\begin{table}[H]
\centering
\begin{tabular}[t]{|c|c|c|c|c|c|}\hline
		\  \ & \ Default (\%) \ & \ Optimized (\%) \ \\ \hline
		\ F-score \ & \ $96.92$ \ & \ $97.92$ \ \\ \hline 
		\ G-mean \ & \ $96.46$ \ & \ $97.63$ \ \\ \hline
		\ Precision \ & \ $95.49$ \ & \ $90.01$ \ \\ \hline 
		\ Sensitivity/Recall \ & \ $98.42$ \ & \ $97.59$ \ \\ \hline 
		\ Specificity \ & \ $95.35$ \ & \ $97.46$ \ \\ \hline
		\ Accuracy \ & \ $96.46$ \ & \ $97.59$ \ \\ \hline 
\end{tabular}
\caption{Column $1$ represents $6$ other performance metrics. Columns $2$ and $3$ show the NDCG score between each of these metrics and the AUC score when ranking $15$ classifiers by their median performance in default and optimized settings, respectively. Median performance of classifiers is similar regardless of which metric to use.} \label{table_ndcg}
\end{table}

\vspace{-1.5cm}
\subsubsection{Performance Variation across Classifiers}
\begin{figure}[H]
\centering
\includegraphics[width=0.85\linewidth]{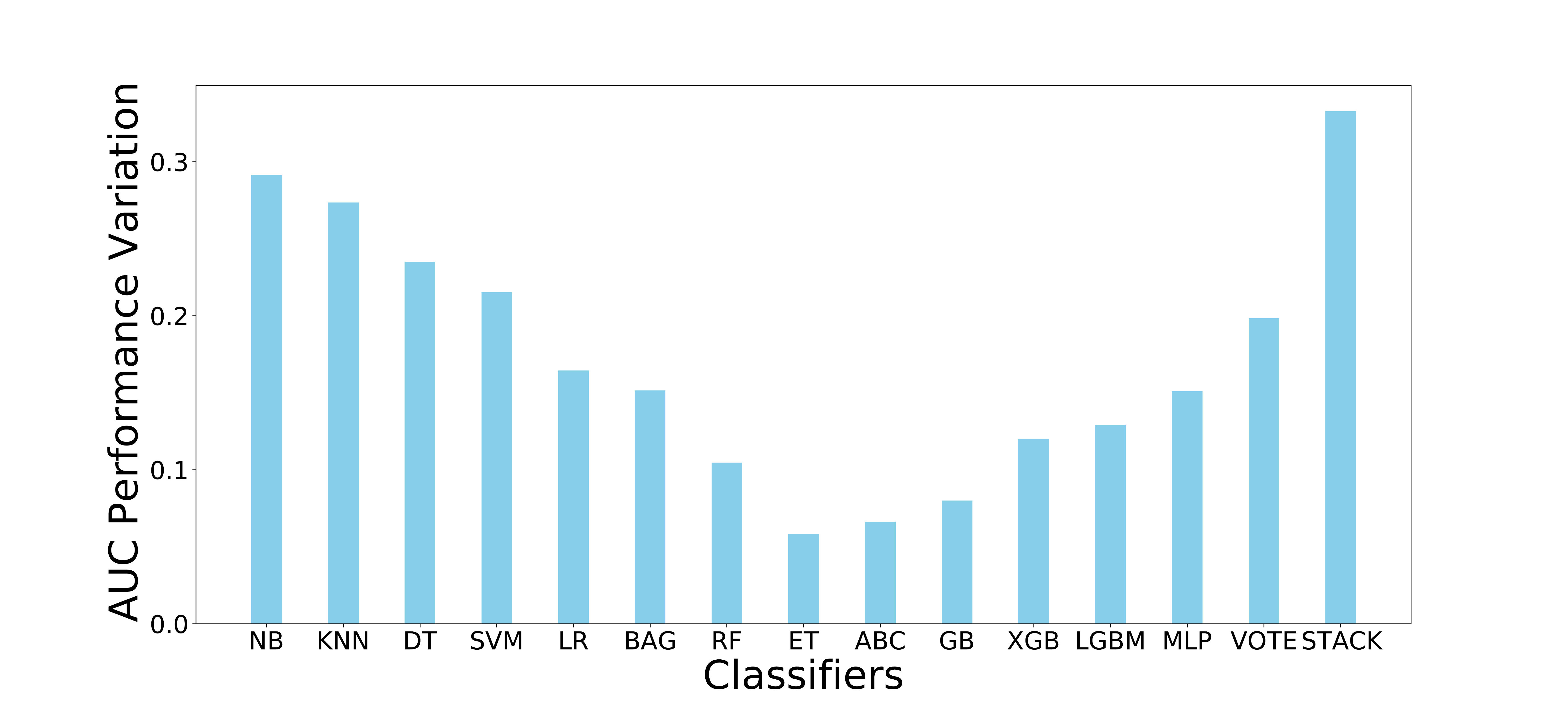}
\caption{Performance variation comparison over all classifiers. The order of classifiers listed on the $x$-axis is based on increasing model complexity. ‘Too simple' and ‘too complex' classifiers result in larger performance variation. The performance variation of classifiers with ‘appropriate' complexity is relatively small.} \label{fig_ccvar}
\end{figure}

We measure the performance variation for each classifier in consideration of all other dimensions, i.e., MV, LB, SCALE, FS, and HT. For each classifier, we get a range of AUC scores. The size of the range determines the extent of performance variation. Fig.~\ref{fig_ccvar} shows the performance variation in the AUC score of all classifiers. The order of listed classifiers on the $x$-axis is based on increasing model complexity, which is measured by classifier training time with default settings.
The complexity of classifiers and performance variation demonstrates an evident ‘U-shaped' relationship. When the classifier is ‘too simple', its performance variation is relatively large. When the complexity of the classifier is ‘appropriate', 
the performance variation is relatively small. If the classifier becomes ‘too complex', it is also at the risk of larger performance variation. Therefore, classifiers with ‘appropriate' complexity are more stable, with smaller changes in performance, while  ‘too simple' or ‘too complex' classifiers are relatively unstable with larger changes in performance in general.

In addition to AUC, the same metrics as above were used to validate the performance variation of all of the classifiers. We use the range (difference between maximum and minimum scores) to rank classifiers in consideration of MV, LB, SCALE, FS, and HT. Then, NDCG is used to compare classifier rankings between each of these metrics and the AUC score. Table~\ref{table_ndcgvar} displays detailed relevance scores. The result suggests that other metrics show a similar ‘U-shaped' relationship between classifier complexity and performance variation as the AUC score. When predicting MOF, it is inappropriate for clinical practitioners to choose ‘too simple' and ‘too complex' classifiers since they may run the risk of underfitting and overfitting, respectively. 
\vspace{-.6cm}
\vspace{-1cm}
\begin{table}[H]
\centering
\setlength{\tabcolsep}{.4pt}
\begin{tabular}[t]{|c|c|c|c|c|c|c|}\hline
		\  \ & \ F-score \ & \ G-mean \ & \ Precision \ & \ \makecell{Sensitivity/ \\Recall} \ & \ Specificity \ & \ Accuracy \ \\ \hline
		\ Relevance (\%) \ & $93.15$ & $94.98$ & $94.37$ & $93.24$ & $93.13$ & $93.77$ \\ \hline
\end{tabular}
\caption{NDCG score between each of $6$ other performance metrics and the AUC score in terms of classifier complexity and performance variation. Different metrics show a similar ‘U-shaped' relationship.} \label{table_ndcgvar}
\end{table}

\vspace{-1.5cm}
\section{Discussion}\label{discussion}
We have provided a timely MOF prediction using early lab measurements (hour 0), patients’ demographic and illness information. Our study quantitatively analyzes the performance via the AUC score in consideration of a wide range of ML configurations for MOF prediction, with a focus on the correlations among configuration complexity, predicted performance, and performance variation. Our results indicate that choosing the correct classifier is the most crucial step that has the 
largest impact (performance and variation) on the outcome. More complex classifiers including ensemble methods can provide better default/optimized performance, but may also lead to larger performance degradation, without careful selection. Clearly, more MOF data is needed to provide a more general conclusion. Our work can potentially serve as a practical guide for ML practitioners whenever they conduct data analysis in healthcare and medical fields.

\section{Acknowledgments}
This work was funded by the National Institutes for Health (NIH) grant
NIH R01 - HL149670.

%
%
%
%
{}
\end{document}